# Gait-Adaptive Navigation and Human Searching in field with Cyborg Insect

Phuoc Thanh Tran-Ngoc[1]†, Huu Duoc Nguyen[1]†, Duc Long Le[1]†, Rui Li[1], Bing Sheng Chong[1], and Hirotaka Sato[1*]

[1]School of Mechanical & Aerospace Engineering, Nanyang Technological University; 50 Nanyang Avenue, 639798, Singapore.

*Corresponding author. Email: hirosato@ntu.edu.sg.
†These authors contributed equally to this work.

**Abstract:** This study focuses on improving the ability of cyborg insect to navigate autonomously during search and rescue missions in outdoor environments. We propose an algorithm that leverages data from an IMU to calculate orientation and position based on insect's walking gait. These computed factors serve as essential feedback channels across 3 phases of our exploration. Our method functions without relying on external systems. The results of our trials, carried out in both indoor ($4.8 \times 6.6$ m$^2$) and outdoor ($3.5 \times 6.0$ m$^2$) settings, show that the cyborg insect is capable of seeking a human without knowing the human's position. This exploration strategy would help to bring terrestrial cyborg insect closer to practical application in real-life Search and Rescue (SAR) missions.

# INTRODUCTION

Terrestrial cyborg insects are developed to serve complex terrain navigations, e.g., post-disaster search-and-rescue missions [1-3]. These hybrid systems combine ambulatory insects as mobile platform and miniature electronic backpacks mounted on their bodies (Fig. 1a). These systems could be controlled to perform various tasks, e.g., autonomous navigation or human detection [2,4]. Their locomotion is regulated by electrically stimulating the insects' muscular/ sensory/ neural systems [4-8]. The power consumed for this locomotion control is negligible (i.e., a few 0.1 mW) [5,9,10], reserving the energy resource for other essential tasks, e.g., localization. Importantly, these systems could exploit vital features of the insects, e.g., their skillful and robust locomotion or the vast collection of natural sensors/receptors [11,12].

The navigation of these terrestrial cyborg insects is progressing significantly. Initial path-follow navigations were demonstrated manually. Automatic systems were then introduced to support studies on the insects' behaviors, display functionalities of the backpacks, or carry out simple demonstrations (e.g., miniature object transportation) [4,13,14]. Practical applications were also studied, e.g., autonomous navigation in unknown environments [2]. Instead of employing obstacle detection sensors (e.g., camera [15]), a navigation program named Predictive Feedback incorporated the insects' innate obstacle-negotiation abilities and artificial control rules to direct them in unknown and obstructed terrains [2]. An infrared (IR) image-based human detection algorithm was also introduced for these cyborg insects [2]. With these achievements, one of the next tasks is to develop an exploration strategy allowing these cyborg insects to proactively navigate in unknown environments while attempting to locate and classify sources of interest, e.g., humans. A simple demonstration of such exploration was attained with acoustic sources [16]. However, the operation range was short, ~1 meter, and the source localization process was not executed onboard, thus wasting power for data transmission and lacking practicability.

In general, an exploration strategy could be divided into three phases, categorized as information objective (Phase I), control/navigation objective (Phase II), and classification objective (Phase III) [17]. In Phase I, the cyborg insect would promptly acquire knowledge about the assigned environment. In Phase II, such knowledge would be used to estimate the searching target's position, which the cyborg insect would navigate toward to gain more information about the target. Once it is sufficiently close to the target, i.e., enough information is collected, the cyborg insect could reliably proceed with the classification objective in Phase III.

Existing literature provides the means to perform Phase III with the IR image-based human detection algorithm [2,18]. The IR image could also be used to execute Phase II. E.g., blob detection techniques in image processing could implement the image's thermal information to locate hot regions inside it [19,20]. These regions could potentially contain the spatial knowledge about the target, thus allowing its position to be estimated [20]. Existing robotic studies report numerous source localization techniques to seek such thermal information promptly, thus supporting the cyborg insect to carry out Phase I efficiently. Random and gradient-based searches are some examples [21,22]. Despite their simplicity, random searches were reported to locate unknown targets effectively [21]. These techniques would also favor the hybrid systems using the IR camera as the main sensor, allowing the lack of environmental gradient measure.

For the purpose of autonomous navigation, accurate knowledge of the position and orientation of the cyborg insect is essential. Traditionally, this information has been obtained through external tracking systems like VICON, which utilize cameras surrounding the subject to determine its location and orientation [2,23]. Such systems are well-suited for indoor environments but fall short in outdoor settings, particularly in the complex and unpredictable terrains encountered during real-life Urban Search and Rescue (USAR) missions. Therefore, the need arises for an onboard localization solution capable of operating independently of external infrastructure. Localization amidst the rubble is crucial not only for navigation but also for accurately reporting the positions of discovered victims. Yet, the development of a suitable positioning system for cyborg insects faces significant hurdles, including restrictions on size, power consumption, and computational capacity. While effective outdoors, global Positioning systems (GPS) are ineffective indoors [24,25] and, like traditional Indoor Positioning Systems (IPS) that rely on cameras for navigation, struggle in the low-visibility conditions of collapsed structures [26-28]. Given that the payload of an insect platform usually matches its own weight—a matter of several grams, primarily occupied by the controller IC and battery—implementing a bulky and heavy localization system like LIDAR is out of the question [29-31]. An alternative approach involves utilizing lightweight, low-power microelectromechanical systems (MEMS) and inertial measurement units (IMUs). However, most of the low power IMU positioning system are designed for pedestrian dead reckoning, which require specific characteristics in human walking gait with separated stance phase and swing phase [32-34]. This cannot be applied to the walking gait of the insect. A recent attempt to use machine learning for insect localization by Cole et al. is still at the post-processing positioning phase which is not ready for on-board application [35].

Herein, this study presents a feasible exploration strategy for terrestrial cyborg insects developed from the above arguments (Fig. 1). The cyborg insect serving this study was made of hissing cockroaches (*Gromphadorhina portentosa*, Fig. 1a) and equipped with the IR image-based human detection algorithm reported in [2]. This algorithm was employed to execute Phase III of the exploration strategy. A blob detection technique was combined with the Predictive Feedback navigation program of [2] to form a Thermal Source-Based Navigation Algorithm used for Phase II. Experiments were set up to evaluate this algorithm's performance and determine the criteria to switch between the three phases. The selection of a suitable stochastic exploration technique used in Phase I was done via a Robot Operating System (ROS) stimulation environment, emulating the natural & controlled motions of terrestrial insects. All three phases were implemented onboard for a demonstration in an unknown environment, verifying their practicability (Supplementary Video 1). Furthermore, this study introduces an algorithm to estimate the position of the insect using a low-power MEMS IMU, which capitalized on the insect body vibration during locomotion. The variance in the acceleration data, reflecting the frequency of the insect's steps and its moving speed, was directly utilized to estimate the insect's linear speed, and the insect's position was calculated through time integration. This IMU-based positioning system supports the exploration strategy for terrestrial cyborg insects capable of operating outdoors (Supplementary Video 2). Limitations and future works were also discussed. The exploration strategy, coupled with the IMU-based positioning method, represents one of the first solutions to meet all hardware requirements while providing adequate accuracy for insect localization. This advancement brings the cyborg insect closer to practical application in real-life Search and Rescue (SAR) missions.

## RESULTS AND DISCUSSION

### *Localization methods of cyborg insect with IMU (Inertial Measurement Unit)*

Based on the results from these experiments, we can prove that our exploration strategy can help the cyborg insects autonomously search around to find the human. However, the navigation method that is implemented on board needs to use the position and orientation of the cyborg insect from the external optical tracking system as the input to navigate the hybrid system in all of its phases. This localization system will eliminate the application of our cyborg insect. Thus, we try to replace the external optical tracking system with an on-board localization system to help the cyborg insect can be operated in real search and rescue scenario condition.

As the cyborg insect targets search-and-rescue application, localization techniques that employed Global Positioning System (GPS) is not suitable due to the potential absence of GPS signal under the rubble. Vision-based methods or range-finder based methods such as LIDAR are not applicable in the dust-filled rubble environment [36-38]. Inertial navigation system using Inertial Measurement Unit (IMU) appear to be the most reasonable method for under-rubble localization. It is well known that MEMS IMU contains noise that prevent the velocity and position of the IMU being obtained by integration method. The size and the sensitivity of IMU and the control backpack also restrict the computation algorithm which can be used for the localization. Such algorithms like step counting [34,39] is not applicable with the reading provided when the backpack is mounted on the back of the insect. However, instead of directly counting step and estimating step length, a simplified method was proposed for positioning calculation which is suitable for insect with walking gait. During insect locomotion, the body of the cockroach shook with similar frequency of the walking gaits, which is from 3 to 9 Hz [40]. The acceleration recorded from natural walking state of the cockroach has also reflected the tendency with a majority of signal frequencies are below 10 Hz (Fig. 2).

Although absolute acceleration magnitude was not reliable, change in acceleration could be used to demonstrate the body shaking in cockroach as this movement was visible to the accelerometer. As cockroach movement is induced by the walking gait, linear speed of the cockroach is dependent on the step-frequency or in another way, body shaking behavior. Hence, an algorithm utilizing acceleration variance to estimate the speed as well as the position of the cyborg insect is proposed:

$$V_{linear} = K * Var(Acceleration)$$

$$\begin{bmatrix} Vx \\ Vy \end{bmatrix} = V_{linear} * \begin{bmatrix} Ox \\ Oy \end{bmatrix} \text{ in 2D tracking or } \begin{bmatrix} Vx \\ Vy \\ Vz \end{bmatrix} = V_{linear} * \begin{bmatrix} Ox \\ Oy \\ Oz \end{bmatrix} \text{ in 3D tracking}$$

$$\begin{bmatrix} X \\ Y \end{bmatrix} = \int \begin{bmatrix} Vx \\ Vy \end{bmatrix} dt \text{ in 2D tracking or } \begin{bmatrix} X \\ Y \\ Z \end{bmatrix} = \int \begin{bmatrix} Vx \\ Vy \\ Vz \end{bmatrix} dt \text{ in 3D tracking}$$

where Acceleration is the gravity-subtracted acceleration recorded by IMU $V_{linear}$ is the linear speed of the insect-machine robot K is the gain factor, which depends only on the insect-machine configuration and nature properties [Vx Vy Vz], [Ox Oy Oz] and [X Y Z] are velocity, orientation, and position vector of the insect-machine system in global coordinate respectively. The algorithm directly estimates the system speed at discrete point in the operation instead of the acceleration integration; hence, the effect of velocity random walk and drifting can be avoided or reduced. The gain K will only need to be calibrated once for each system-environment combination by letting the cockroach run freely for 5 seconds. The position of the cockroach calculated on-board from the IMU data was verified to be accurate by comparing it with the position recorded by the external optical tracking system (Fig. 3). Throughout the test duration, the error of the IMU-based estimated position remained less than 1m and propagated at the scale of approximately 5% of the travelled distance.

The error accumulation could be the results of misalignment between the IMU orientation and the actual moving direction of the cockroaches caused by the posture change as well as the false movement detection. The cockroaches tend to lift up the body to examine when it encounters an obstacle and attempt to climb over obstacle as a negotiation method which leads to the rise in the pitch angle of the IMU-mounted PCB. The algorithm registered it as a walking movement which caused an error in positioning. Even though the climbing attempt would be cancelled due to the height of the obstacles, errors accumulated would continuously increase, especially in the positive Z direction.

*Performance of the Thermal Source-Based Navigation Algorithm*
*With the external optical tracking system*
As previously discussed, our human detection algorithm is effective when enough information about the subject is present in the thermal image obtained from the camera. The current IR camera allows for high-accuracy identification of people in the range of 0.5m to 1.5m. However, as the distance from the camera to the subject increases, the model's reliability decreases. As an alternative, an alternative method can be implemented by identifying the direction of the heat source in the camera and then controlling a cyborg insect to approach the heat source for increased subject information. Besides, the algorithm should not be too complicated so that the model can be operated onboard and in real-time.

The Thermal Source-Based Navigation Algorithm successfully navigated the insect towards the target object in 28 out of 30 trials (93.3%, N=3 insects). The average navigation time was 111.5 s and the average linear speed of the insect was 3.7 cm/s. The position of the target was generally estimated three times per navigation. The average distance of the insect to the target was reduced from 4 m to approximately 2.5 m and 1.0 m after arriving at the first and second estimated destinations, indicating a successful position estimation (Fig. 4a). The thermal information collected by the hybrid systems also increased as the navigation progressed, demonstrating the potential of the algorithm for use in Phase II.

The two failed trials were caused by the cyborg insect exceeding the limited time before reaching the oven. The problem occurred because the cyborg insect was stimulated, causing it to turn quickly when the IR camera on the backpack captured the oven. As a result, the cyborg insect's orientation at this time was different from its orientation when the thermal image was taken. Since the estimated target of the hot object is determined based on a combination of the cyborg insect's direction and the location of the hot object within the thermal image, the first target was not in the direction of the oven. Thus, after the cyborg insect reached the first target, its direction was not facing the oven, preventing the infrared (IR) camera mounted on its backpack from capturing a new thermal image of the oven. Then, the algorithm was not activated to provide the cockroach with the second target.

As the target's position was heuristically searched via the approaching step L, there were overshoots causing the third estimated destination to lie behind the target (27/28 trials, Fig. 4b). The approaching step L could be dialed down or adaptively adjusted with the thermal information to overcome this issue. While these solutions are worth investigating, they might not be power efficient due to the high operation rate of the IR camera. An alternate solution would be promptly switching from Phase II to Phase III when the thermal information is sufficient for a reliable classification process. As introduced, such a process would require the IR image to contain at least 4.9% of thermal information within the range of 28 °C to 38 °C. Implementing this as a phase-switching criterion, Phase III could promptly start after the insect arrived at the second destination (in 27/28 trials, ~96.4%).

*Utilizing onboard IMU-based localization system*

In the previous section, the position and orientation data from external optical tracking system were utilized to navigate a cyborg insect towards a heat source, identified as the oven. The distance between the oven and the cyborg insect, along with the predetermined approaching step L, guided the process. The blob detection algorithm was activated only three times to determine the direction of the heat source and to set temporary targets. This method was designed to minimize processing delays. However, there was a chance of misdirection because of potential errors in determining the heat source's direction when the cyborg insect rotated. Furthermore, the fixed step size L often resulted in overshooting the target.

To address these issues, our Thermal Source-Based Navigation Algorithm was refined by developing a new approach that leverages the IMU's yaw angle for enhanced orientation correction. Consequently, the blob detection was enabled to capture the heat source as the cyborg insect walked during phase II. Whenever the heat source moves beyond the camera's FOV, the algorithm utilizes orientation feedback from the IMU to control the cyborg insect and turn it back to recapture the heat source. This adjustment resulted in a success rate of 100%, an increase from 93.3% achieved with the external optical tracking system. Additionally, the onboard IMU-based localization system caused the average navigation time to extend to 124.7 seconds and the average linear speed to increase to 5.0 cm/s. These statistics show an 11.8% longer navigation time and a 35.1% faster speed compared to the external tracking-based Thermal Source-Based Navigation Algorithm, which had previously recorded an average time of 111.5 seconds and a speed of 3.7 cm/s. The navigation time has lengthened, and the speed has increased due to our upgraded navigation algorithm. This improvement constantly tracks the heat source and leverages the orientation data from IMU to fine-tune the movement of our cyborg insect throughout phase II. Thus, frequent adjustments by electrical stimulation are required to realign the IR camera's field of view with the oven (Fig. 4c).

These improvements underscore the efficacy of the onboard IMU-based localization system into the thermal source-based navigation algorithm, enabling the cyborg insect to navigate more effectively towards heat sources in phase II of our study.

***Enhancing Stochastic Exploration Strategies through IMU Integration***

*Selection of the Stochastic Exploration Strategy*

The stochastic exploration strategies were performed differently. The coverage rate, i.e., the coverage percentage over time, was slowest with the natural walking cyborg insect, perhaps due to its strong wall-following tendency (Figs. 5a, 5c). Fixed Length performed better with a coverage of 61% after 24 hours (Fig. 5b). However, its local search characteristic resulted in a slower coverage rate than the other three strategies, implementing distant search (Figs. 5a, 5c). Despite all having a final coverage of more than 90%, Brownian Walk's coverage rate increased faster than the other two as purely favoring distant explorations. As the local searches (i.e., nearby destinations) in Levy Walk outweighed its remote explorations (i.e., faraway destinations), its coverage rate was the slowest among the three.

However, it slightly outran Uniform Distribution and Brownian Walk for locating the unknown target promptly. This strategy's average searching time was 221 min, up to ~13% faster than the other two (Figure 5d). Like the coverage, all three strategies outperform Fixed Length, as their global exploration ability resulted in a higher chance of early encountering the target.

However, as Levy Walk allowed the cyborg insect to seek around its vicinity before distant jumps, its opportunity to quickly detect the target would be slightly improved further (Fig. 5e). As a trade-off, this local search characteristic might contribute to its failed cases (Fig. 5d). Unlike in other studies [21,41], Levy Walk's superiority in searching for unknown targets was insignificant herein. As being a proposed stochastic exploration strategy for locating scarcely distributed targets in a vast space [42], its performance in a bounced environment (like this study) might decline

As shown, Levy Walk, Brownian Walk, and Uniform Distribution could all be potentially used for Phase I. Each was slightly better than the others in either coverage rate, searching time, or success rate. It is worth conducting further simulations and analyses to get a more comprehensive view, e.g., altering the numbers and positions of unknown targets and cyborg insects. Due to its slight edge on the search time, this study implemented Levy Walk as the random search strategy. While following this strategy, the cyborg insect would regularly gather information about the surroundings using its infrared camera.

*Stochastic Exploration Strategy with IMU*

An Inertial Measurement Unit (IMU) has been integrated into the backpack to calculate the rotational and positional data (Fig. 6a). This means that the previous reliance on the external optical tracking system has been eliminated. The experimental area was partitioned into 600 squares, each measuring $10 \times 10$ cm, and was deemed to be covered once crossed by the cyborg insect. During the experiments, three hybrid systems were used in a total of nine trials (N = 3, n = 9). The average coverage reached within the first ten minutes was 22.8%, with a standard deviation of 5.5% (Figure 6c). An essential characteristic of this strategy is the implementation of a Lévy Walk exploration pattern. This approach enables periodic focus on adjacent objectives, alternating with substantial movements toward more distant locations, so optimizing search efficiency by reducing the need to revisit previously explored spaces (Figure 6b). In addition, the system gains advantages from the incorporation of an infrared camera, capable of detecting heat sources at a maximum range of 4.2 meters. This greatly expands the range of operations during the initial phase of exploration (Phase I).

*Design and Demonstration of the Exploration Strategy*

As previously mentioned, the exploration strategy has three distinct phases: Phase I involves implementing the Levy Walk technique, Phase II utilizes Thermal Source-Based Navigation, and Phase III employs the IR image-human detection algorithm. The thermal data from the IR image can be utilized as a parameter for initiating Phase III, which requires a minimum of 4.9% of pixels falling within the temperature range of 28°C to 38°C, approximately equivalent to 50 pixels. This information can also serve as a reference point for the commencement of Phase II. The blob detection technique exhibits a high level of reliability within a range of 4.2 meters or less, with an accuracy exceeding 86.9% (Fig. 7a). Hence, a thermal information level of 0.8% (equivalent to 8 pixels) within this effective working range can serve as the threshold for transitioning from Phase I to Phase II (Fig. 7a).

The exploration technique devised in this work was executed entirely on the onboard device (Fig. 7c). The memory utilized by the blob detection technique, which is an image processing task, and the human detection, which is a machine learning model, was relatively little. It occupied roughly 1% (~23.2 kB) of Flash memory and 64% (~163.5 kB) of SRAM memory. In addition, the onboard IMU-based localization method consumes approximately 0.2% (~4.6 kB) of flash memory and 5% (~12.9 kB) of SRAM. This allows for the inclusion of additional features, such as the filter method. The computational time of blob detection, human detection, and IMU-based localization was approximately 350 ms, 95 ms, and 10 ms, respectively. These processing times are sufficiently fast to maintain real-time processing given the navigated insect's speed of around 5.0 cm/s. In comparison to previous research, this technique has shown enhancements. For example, the range for localizing the seeking target can reach up to 4.2 meters in Phase II, which is a major improvement from the previous range of 1 meter achieved by acoustic information [16]. In addition, the integration of this technology into the device would contribute to a decrease in power consumption during data transmission. For example, the energy required to capture infrared (IR) photos was roughly 24 milliwatts, which is about half the energy necessary to stream the auditory data [16].

The efficacy of the exploration technique was demonstrated in an indoor environment, where the cyborg insect effectively detected and recognized a human without any prior information about their location (Supplementary Video 1). This was accomplished in an enclosed environment with a surrounding temperature of around 25 degrees Celsius. Nevertheless, doing trials outdoors at night presented extra difficulties due to the fluctuating external temperatures ranging from 28°C to 29°C. Additionally, the ambient temperature fluctuations and the heat emitted by lighting and air conditioning units have the potential to cause inaccuracies in the Thermal Source-Based Navigation Algorithm.

To address these challenges, we have improved our approach to better differentiate human heat signatures from other sources of environmental heat. For example, the temperatures of lights and air conditioning systems above the typical range of human body temperatures, which often varies from 29.3 to 34.9 degrees Celsius. To ensure correct phase transitions, we used a temperature range threshold of 29°C to 35°C. In addition, ambient temperature fluctuations add complexity because their range coincides with that of human temperatures. However, the dispersed quality of the air presented a distinct chance for improvement. Through the examination of the distribution of pixel temperatures surrounding the central point of the identified thermal source, we have developed a dependable detection procedure. If the IR image contains more than 25 pixels within the range of human temperatures, with a minimum of 12 of those pixels concentrated in a 5 x 5 pixel area at the center, the algorithm will navigate the cyborg insect towards that target (Fig. 7c). However, this modification required a compromise, resulting in a reduction of our blob detection's effective range from 4.2 meters to 1.8 meters (Fig. 7a).

Occasionally, changes in the surrounding temperature may initially satisfy the criterion for detection but then fail in subsequent checks because of their temporary nature. When the thermal source-based navigation system is unable to identify the thermal source again, the algorithm relies on the IMU's yaw angle data to modify the orientation of the cyborg insect. For instance, if the cyborg insect initially turns to the left and fails to detect the heat source, the algorithm will subsequently command it to turn right in an effort to recapture the thermal source. If a complete 90-degree turn to the right does not find the source, the cyborg insect is controlled to perform a comparable left turn of up to 90 degrees. In the event that the target is not discovered, the algorithm disregards the prior detection and returns to Phase I. This prompts the cyborg insect to resume its search, as shown in Figure 7c.

The outdoor demonstrations, employing the enhanced algorithm, successfully mitigated interference from non-human heat sources such as lights and air conditioning units. Regarding environmental warm air, the system encountered it three times; each instance, although initially meeting detection criteria, vanished in subsequent captures. Consequently, after attempts to recapture these fleeting thermal sources, the algorithm disregarded these detections, returning to Phase I. The strategy persisted until the cyborg insect located a human seated at an intersection, with a proximity of approximately 1.7 meters, thereby demonstrating the efficacy of our adaptive approach in complex environmental conditions. The effectiveness of the improved exploration technique was shown in an outdoor setting, where the combined system successfully explored and identified a human without any prior knowledge of the human's position (Supplementary Video 2).

**Acknowledgments**
The authors thank Mr. Ho Yow Min, Mr. Ying Da Tan, Mr. Terence Goh at KLASS Engineering & Solutions Pte. Ltd, Mr. Cheng Wee Kiang, Mr. Ong Ka Hing, Ms. Rui Huan AW at Home Team Science & Technology Agency (HTX), and distinguished officers at Singapore Civil Defence Force (SCDF) for their helpful comments and advice, Ms. Kerh Geok Hong Wendy, Mr. Roger Tan Kay Chia for their support. A part of this work was supported by KLASS Engineering & Solutions Pte. Ltd (NTU REF 2019-1585).

**Author contributions**
H. S., P.T.T.N., H.D.N., D.L.L. conceived and designed the research. H.D.N., D.L.L., and P.T.T.N. developed hardware and software for the backpack. P.T.T.N., H.D.N., and B.S.C. developed the thermal source-based navigation algorithm. H.D.N. and P.T.T.N. developed the Stochastic algorithm. P.T.T.N., H.D.N. developed the exploration strategy. D.L.L. developed the IMU-based localization. P.T.T.N., R.L. prepare the cyborg insects. P.T.T.N., H.D.N., D.L.L., R.L. conducted the experiment and analysis. P.T.T.N., H.D.N., D.L.L., B.S.C and H.S. wrote and edited the manuscript. H.S. supervised the research. All authors read and edited the paper.

**Competing interests**
The authors declare no competing interests.

**Online content**
Any methods, additional references, extended data, supplementary information, and code are available at XXX

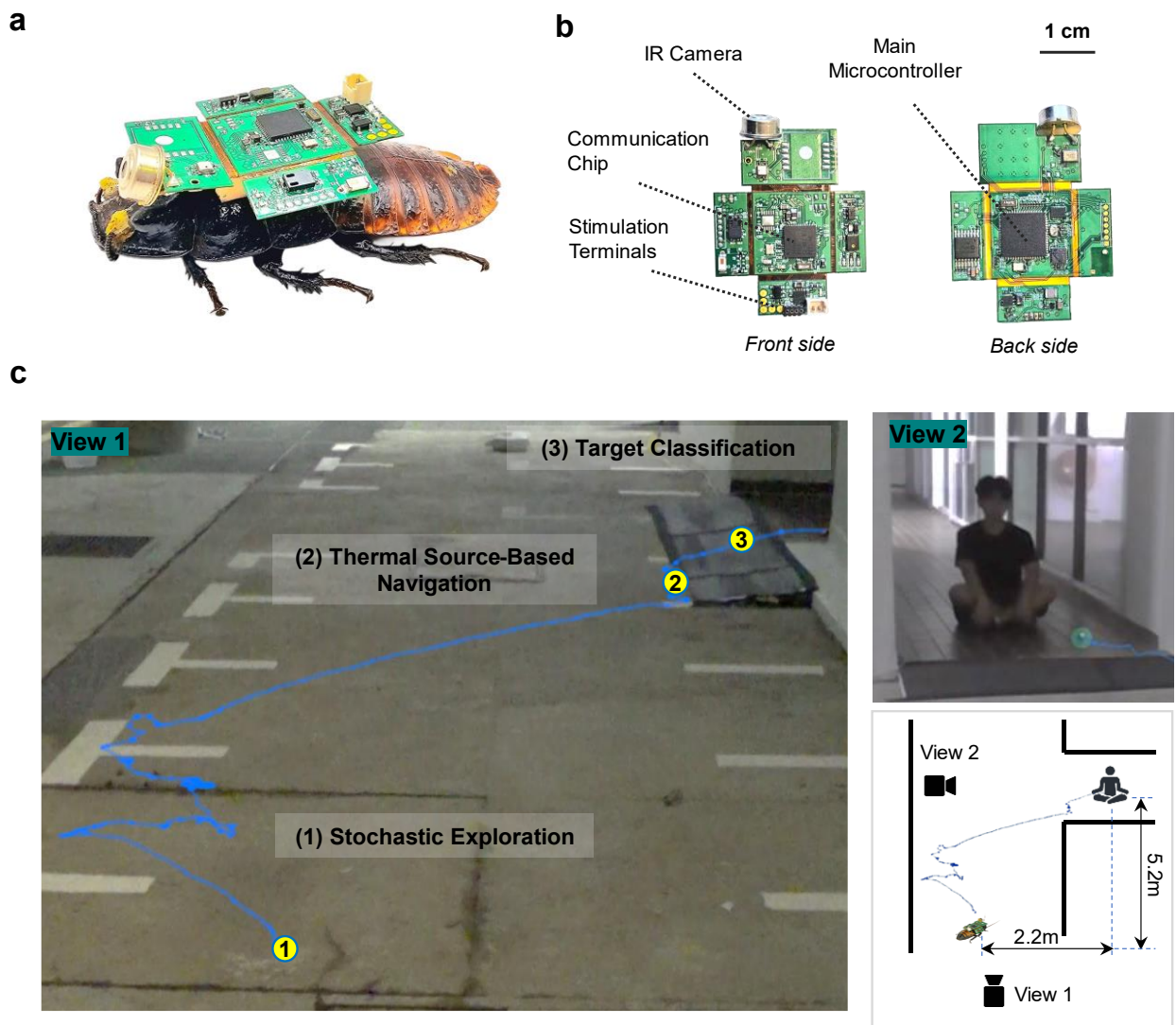

**Fig. 1 | Overview of the cyborg insect and the exploration strategy. a**, The cyborg insect. It is made of a living Madagascar hissing cockroach and an electronic backpack. **b**, The backpack & its essential components. **c**, The process consists of three stages: (1) identifying the objective through a stochastic exploration strategy, (2) approaching it by utilizing thermal information as a guide, and (3) categorizing it using a machine learning (ML) model. The trajectory of the cyborg insect is represented by a blue line, while its position is indicated by a highlighted green circle.

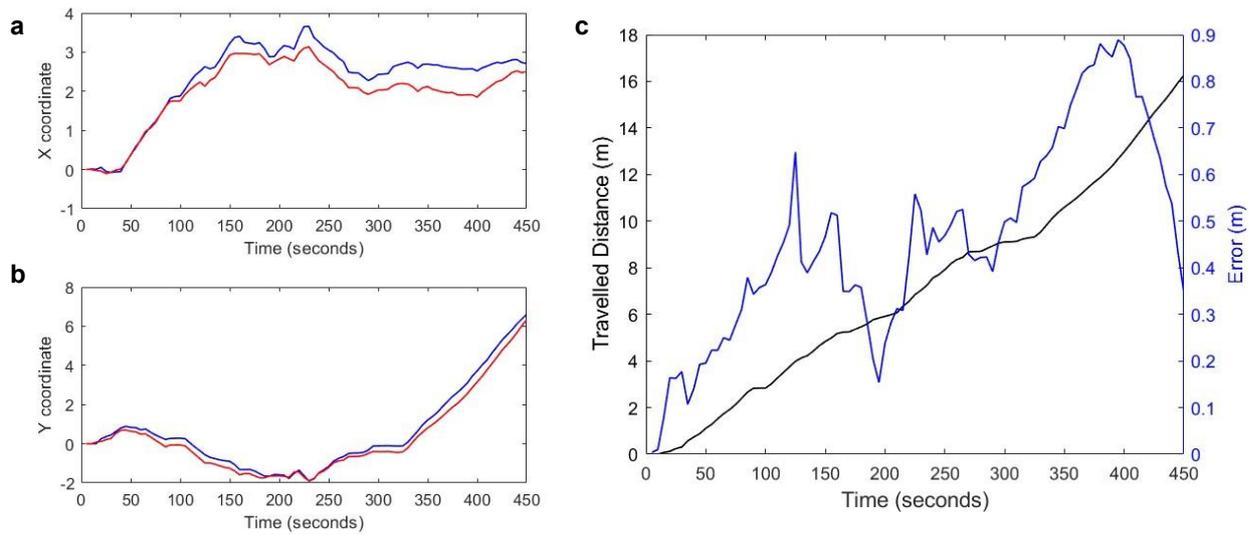

**Fig. 2 | IMU-based localization system performance in 2D**. **a-b**, Position estimation in 2D is based on IMU and compared to the true position provided by an accurate position tracking system. **c**, Error propagation during the cockroach walking. The scale of the error bar is 5% of the traveled distance scale. Estimated positions are shown in blue, while true positions are depicted in red.

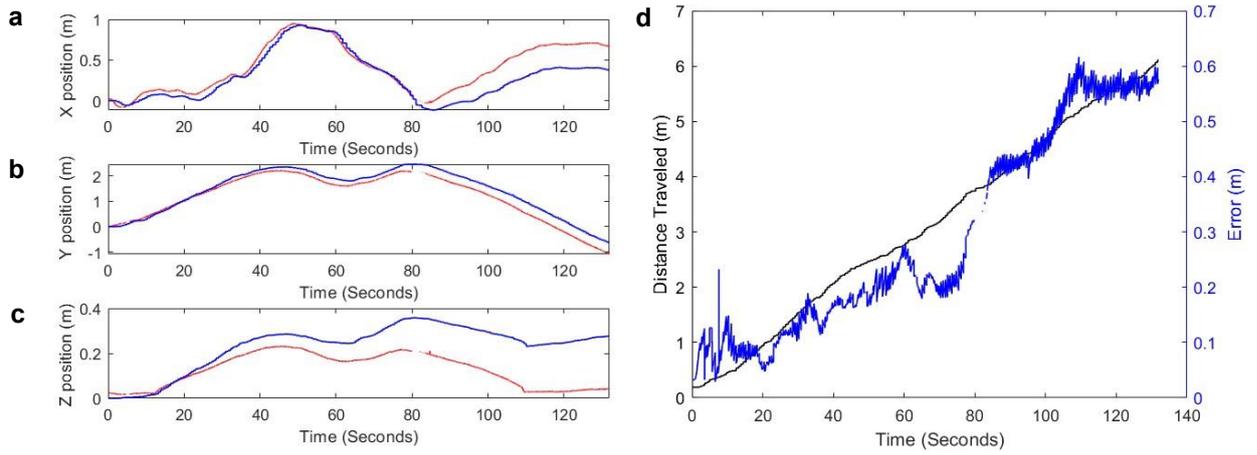

**Fig. 3 | IMU-based localization system performance in 3D**. **a-c**, Position estimation in 3D based on IMU compared to true position provided by the accurate position tracking system. **d**, Error propagation during the cockroach walking. The scale of the error bar is 10% of the travel distance scale, which was significantly higher than the 2D positioning error. The empty data point was due to a marking missing in the external optical tracking system. Estimated positions are shown in blue, while true positions are depicted in red.

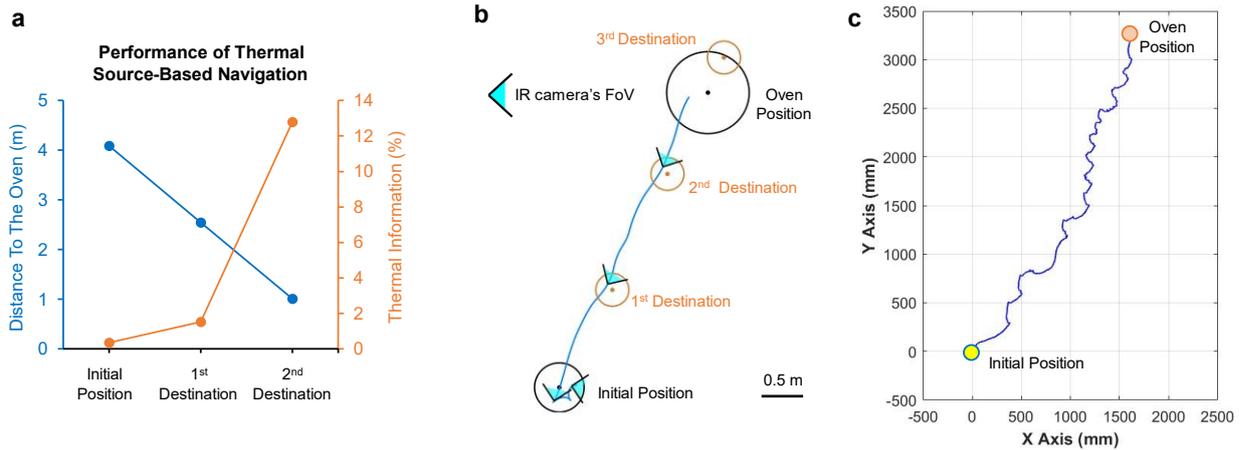

**Fig. 4 | Performance of the Thermal Source-Based Navigation Algorithm. a**, The thermal information's increase (28 °C to 38 °C) and the decrease in the insect's distance to the oven show that it is controlled toward the oven without knowing its position. Data are shown as mean. **b**, A typical navigation trajectory that uses the optical tracking system for localization is the blue curve. After the oven falls into the IR camera's FoV, the algorithm is activated. The orange circles denote the destination arrival regions, their radius is 0.2 m. **c**, A typical navigation trajectory that uses the onboard IMU-based localization system is the blue curve. The algorithm is triggered once the oven enters the field of view (FoV) of the infrared (IR) camera. If the thermal source is not visible in the camera's field of view during movement, the algorithm will utilize the yaw angle from the IMU to regulate the cyborg insect and align its orientation in order to consistently track the thermal source.

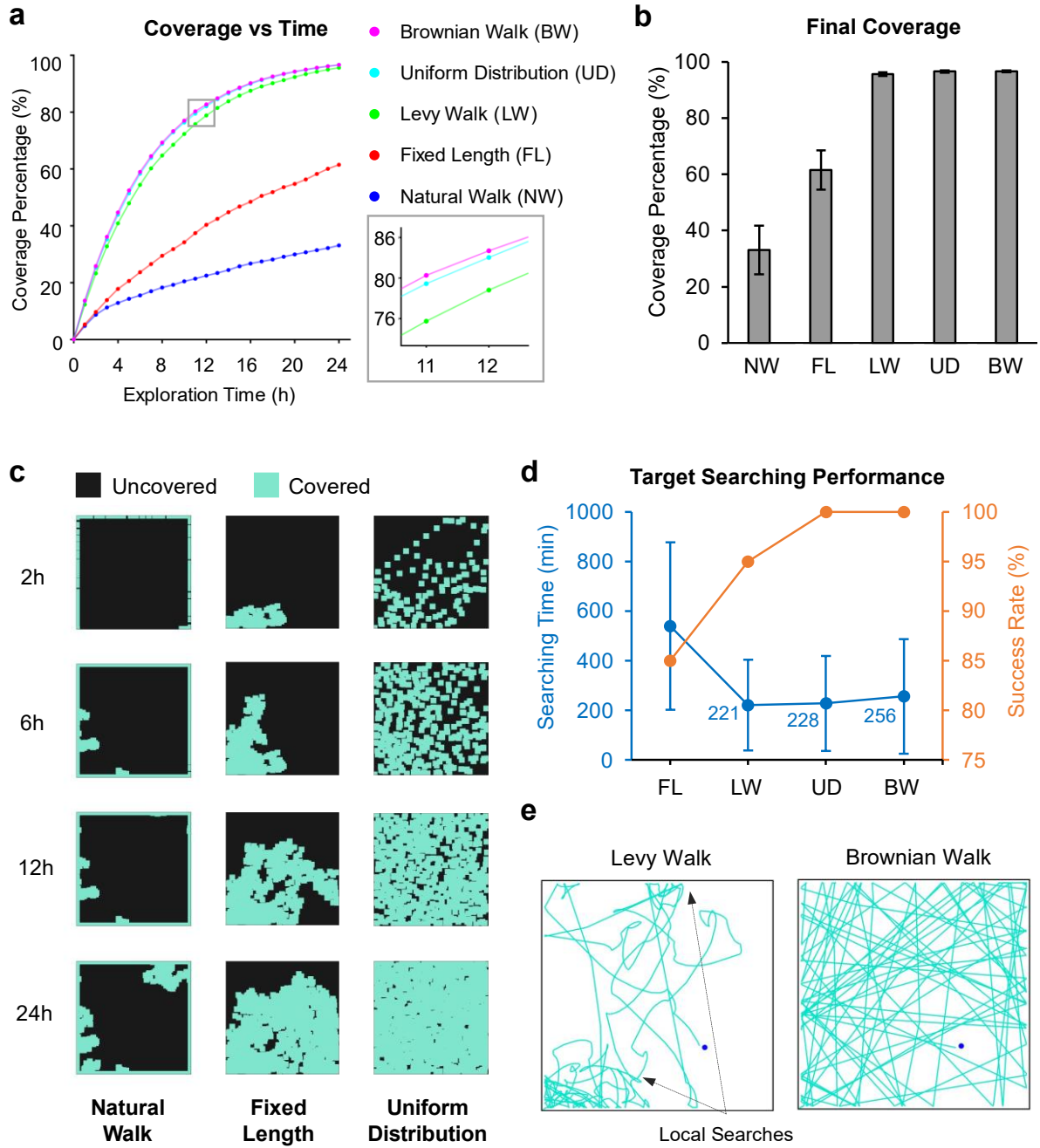

**Fig. 5 | Performance of random search strategies**. **a-c**, Their performance in covering empty space. Brownian Walk, Uniform Distribution, and Levy Walk provide the fastest coverage rates. Under natural walk, the insect spends much time staying near the walls. Fixed Length provides a slow coverage rate as it favors searching local regions. **d-c**, Their performance in searching for an unknown target. Levy Walk edges Brownian and Uniform Distribution via its rations between local and distant searches. **e**, the blue dot and cyan lines denote the target and the robot's trajectory, respectively. Data are shown as mean ± standard deviation.
22

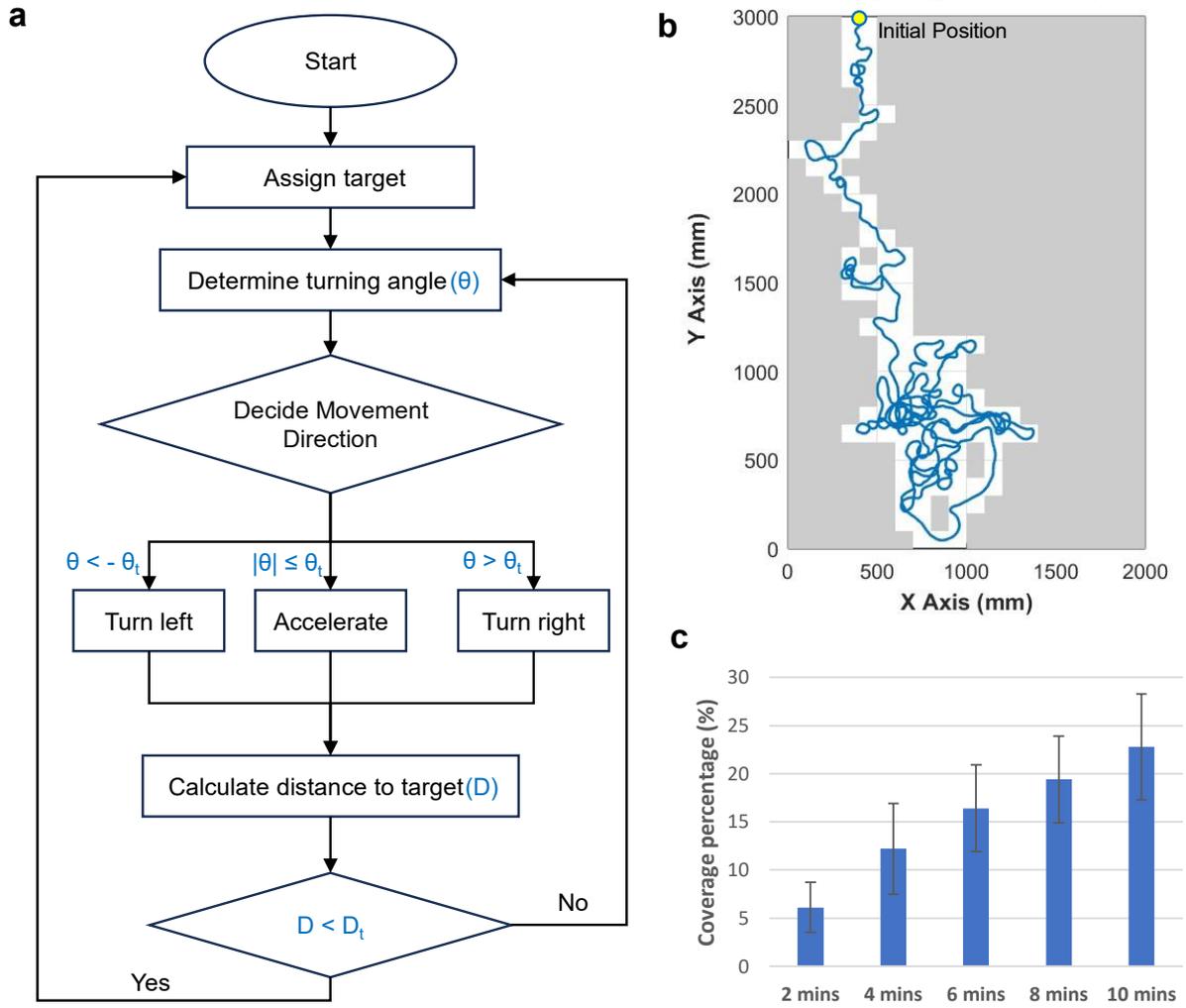

**Fig. 6 | The stochastic exploration strategy with onboard IMU-based localization system in cyborg insects**. **a**, Flowchart of the Stochastic Exploration Strategy. This diagram illustrates the process of navigating the cyborg insect from a starting location to a target generated by a Levy walk. The method calculates the orientation angle (θ) and the distance (D) to the target using the yaw angle and position data obtained from the gait-adaptive IMU. The determination to accelerate, turn left, or turn right is based on comparing the angle θ with the stated threshold angle ($θ_t$) of 20 degrees, and evaluating the distance D against the threshold distance ($D_t$) of 10cm. The process repeats, directing the cyborg insect towards a specific arrival zone described by $D_t$, and then selecting a new target, thereby allowing the cyborg insect to explore the specified area. **b**, The exploration path of the cyborg insect utilizing the Levy walk strategy over a duration of 10 minutes, achieving an area coverage of approximately 18.3%. The yellow dot symbolizes the initial point, while the blue line depicts the path taken by the cyborg insect during its exploration. The white color highlights the area being searched. **c**, Incremental area coverage over 10 minutes, demonstrating the exploration efficiency (N = 3 cyborg insects, n = 9 trials). Data are shown as mean ± standard deviation.

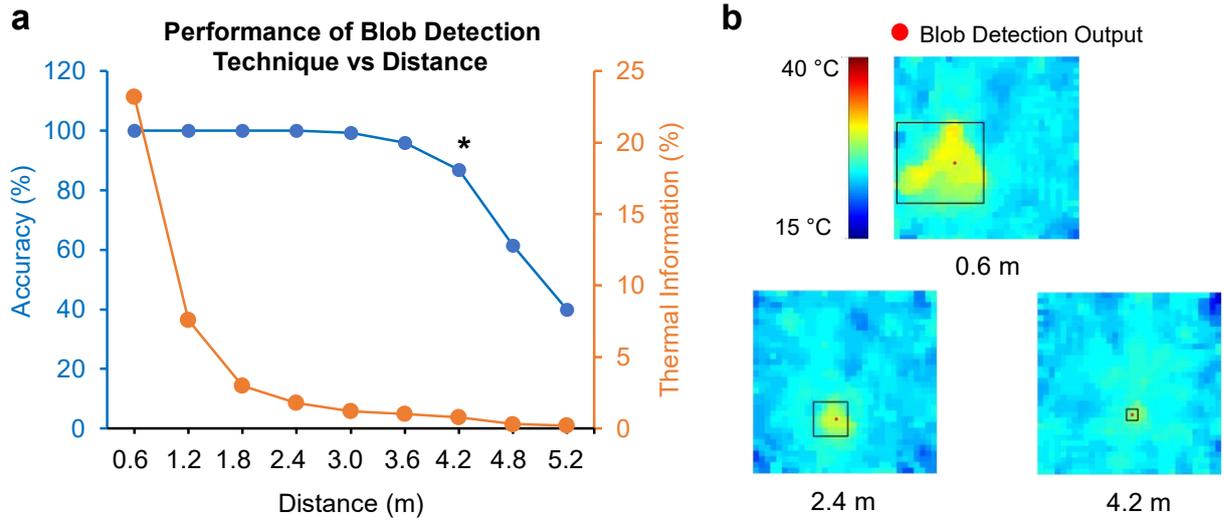

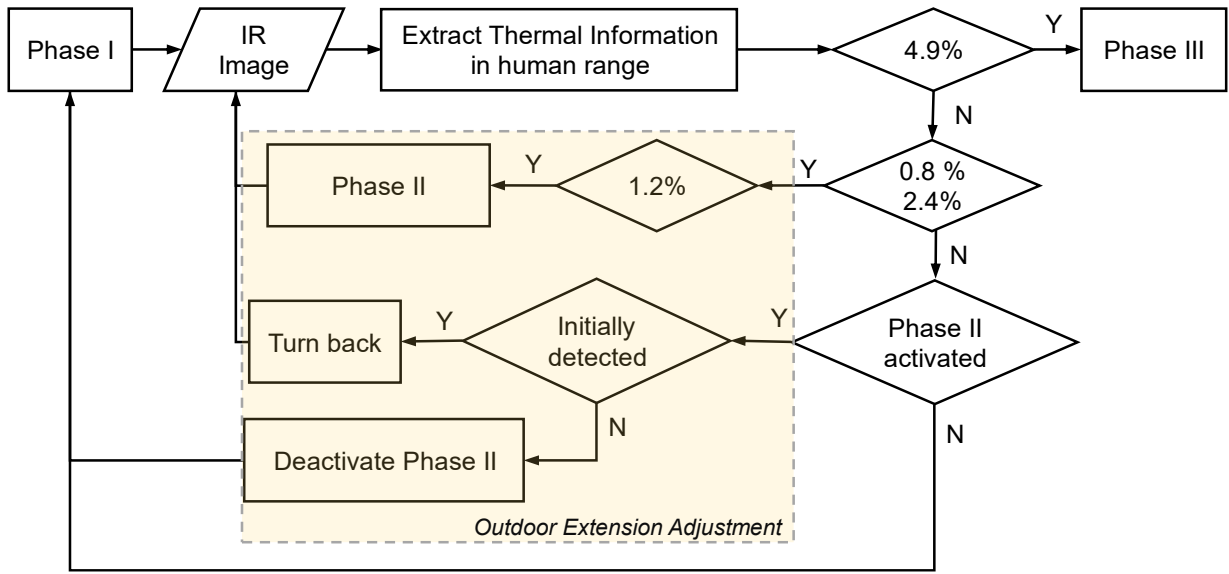

**Fig. 7 | Design of the exploration strategy**. **a**, The blob detection technique's accuracy reduces as the distance between the IR camera and the target increases. 4.2 m (*) could be selected as the maximum distance for acceptable accuracy. This distance's corresponding thermal information (28 °C to 38 °C) is 0.8%. Data are shown as mean. **b**, IR images and their detected blobs (i.e., red dots) at different distances. The square boxes indicating the image's hottest region are used to evaluate the blobs' accuracy. **c**, Flowchart of the exploration strategy for both indoor and outdoor applications. The Indoor Exploration Strategy utilizes the thermal information from the IR image within a specific range to facilitate the transition between three distinct phases. An Outdoor Extension Adjustment module enhances this strategy for outdoor environments by analyzing pixels in a human range near the thermal blob's center. Besides, If a heat source is detected but then lost when the cyborg insect has moved out of the camera's field of view, the algorithm will navigate the cyborg insect and turn back to reattempt capturing the heat source. The algorithm reverts to Phase I: Stochastic Exploration when it continuously fails to find the heat source, forcing a repeated search for thermal sources in outdoor areas.

**METHODS**

Cyborg insect & Its Onboard Human Detection Algorithm

This study reassembled the Cyborg insects used in [2] (Figs. 1a-1b). Madagascar hissing cockroaches (*Gromphadorhina portentosa*, ~6 g, 6 cm) were selected as the insect platform (Fig. 1a). The insect's locomotion was controlled via the electrical stimulation of its cerci [2]. The stimulation of individual cercus caused opposite turns, whereas simultaneously stimulating two cerci resulted in accelerations. The backpack was inherited from [2] (Fig. 1b). It featured a wireless communication ability (CC1352, TI, 48 MHz, 352 kB of Flash, 8 kB of SRAM), an IR camera of $32 \times 32$ pixels ($90° \times 90°$ FoV, Heimann Sensor GmbH), and four stimulation channels (Fig. 1b). The Predictive Feedback Navigation Program and IR image-based human detection algorithm were embedded into the onboard microcontroller (MSP432P4011, TI, 48 MHz, 2 MB of Flash, 256 kB of SRAM) [2]. Despite employing a low-resolution IR camera, the human detection algorithm was reportedly able to classify between humans and everyday thermal items (e.g., laptops, ovens) with the success of up to 90% within the range from 0.5 to 1.5 m, equivalent to at least 4.9% of the image contained information about human temperature, i.e., 28 °C to 38° C [2].

Design of the Thermal Source-Based Navigation Algorithm

Phase II, this algorithm would estimate the target's position, then proceed with navigation (Extended Data Fig. 1). The navigation was executed with the Predictive Feedback program. The target's position was estimated as follows. First, a blob detection technique would locate the target's coordinate ($u$, $v$) inside the IR image (Extended Data Fig. 1a) [19]. After using a Median filter ($3 \times 3$), this technique smoothed the image with a Gaussian filter and then nominated its hot spots (potentially representing the target) with a Laplacian filter ($3 \times 3$) [19] (Extended Data Fig. 1a). Three Gaussian filters were used, including $33 \times 33$ ($\sigma = 5$), $27 \times 27$ ($\sigma$ = 4), and 21 × 21 ($\sigma$ = 3). The coordinate ($u$, $v$) was selected between the nominated spots with the lowest Laplacian output.

External optical tracking system for Thermal Source Navigation

The coordinate was then mapped to α (deg), the target's angular position in the IR camera's horizontal FoV (1) (Extended Data Fig. 1b). Finally, the angle α was used with the insect's location ($X_I$, $Y_I$, $\theta_I$) and the approaching step $L$ (m) to estimate the target's position ($X_T$, $Y_T$) (2)-(4) (Extended Data Fig. 1b). $L$ was tuned to 1.5 m based on the reliable working range of the human detection algorithm [2].

$$\alpha = (u - 1) \times 90°/31, u \in [1\ 32] \quad (1)$$

$$\Upsilon = \theta_I - 45° + \alpha \quad (2)$$

$$X_T = X_I + L \times \cos(\Upsilon) \quad (3)$$

$$Y_T = Y_I + L \times \sin(\Upsilon) \quad (4)$$

Once the cyborg insect arrived at the estimated destination, a new IR image would be captured for either a new estimation or the classification (Phase III). Upon its arrival, as the insect's orientation was not a control output [2], the target might be out of the IR camera's FoV, thus preventing the new estimation. Under such a scenario, an additional destination would be issued with the approaching step of Δ$L$ (m). The insect would then be navigated between these two destinations to alter its orientation. The step of Δ$L$ was selected as 0.2 m.

Onboard IMU-based localization system for Thermal Source Navigation

Based on the target's coordinates (u,v) inside the thermal image, the system's navigation decisions are made according to the blob center's position on the u-axis within the thermal image. For example, the system turns left if the position falls within 1 to 10 pixels, turns right if within 23 to 32 pixels, and accelerates if within 11 to 22 pixels (Extended Data Fig. 1d). By utilizing the yaw angle ($\psi$) from the gait-adaptive IMU, we calculate the rotating angle after the system is navigated based on the blob detection results. This rotating angle is compared with the target's angular position (α) when the heat source is outside the IR FoV. Depending on the difference between the rotating angle and α, the algorithm adjusts the system's orientation, enabling the IR camera to capture the heat source during the movement of the cyborg insect.

Evaluation of the Thermal Source-Based Navigation Algorithm

Two experiments were conducted. In the first experiment, the algorithm was examined using a batch of 3 cyborg insects navigated by the algorithm to approach a thermal target i.e., an oven (28 – 50 °C), placed 4 m from their initial location (Extended Data Fig. 1c). The oven's position was unknown to the algorithm. Successful navigations were recorded when the insect arrived at the target, i.e., less than 0.5 m away from the oven, within 180 s, otherwise, a failed navigation was recorded. The experimental arena was $4.8 \times 6.6$ m$^2$. The ambient temperature was set as 25 °C. The cyborg insect's initial orientation was randomized. The target's position was estimated when the IR image contained more than 0.2% thermal information within 28 °C – 38 °C. For external localization, the insect's position was monitored via an external optical tracking system (VICON®, 10 cameras, 100 fps, Extended Data Fig. 1c). 10 trials were conducted for each cyborg insect (n = 30). For self-localization, the cyborg insect's orientation was determined through an integrated IMU (100 fps, Extended Data Fig. 1e). This part of the experiment was conducted over 9 trials (n = 9).

The second experiment was to set the switching criteria between the three phases. Several sets of IR images containing human were collected at different distances (Figs. 7a-7b). The IR camera was situated on the floor at the height of 20 mm, approximating insect's height [2]. The ambient temperature was set at 25 °C. Inside the camera's FoV, a human was presented at different positions (Fig. 7b). The distance between the human and the camera was changed from 0.6 m to 5.4 m with a step of 0.6 m. At each distance, at least 200 images were captured and processed to extract the human coordinate inside them using the blob detection technique (Fig. 7b). If this coordinate was inside the hottest region of the image, the blob detection was judged as performing successfully. This region was constructed from the segmentation algorithm proposed in [18].

Development of the ROS Simulation Environment

As introduced, a ROS simulation environment was built. ARGoS and its foot-bot model were selected as the engine and platform to simulate (Extended Data Fig. 2c) [43]. Three motions were programmed for the robot: forward motion, directional turn, and stationary, representing the 2D model of a living insect's locomotion (*Gromphadorhina portentosa, Blattella germanica*) [44,45]. The robot emulated two statuses of the insect, natural walking or under navigation. The insect's linear and angular speeds depended upon each status (i.e., $v$, $\omega$, Extended Data Table 1). The insect's natural walk could be modeled as a finite-state machine, switching between walking in open space and following walls (Extended Data Fig. 2b, Extended Data Table 1) [44,45].

- In an open space, the insect would switch between moving and standing still with the probability of $P_{stop}$ [45]. The stopping duration, $T_{stop}$ (s), was drawn from a normal distribution (Extended Data Table 1) [45].

- Each move contained a straight walk of $L$ cm, followed by an instantaneous turn of $\alpha$ deg, drawn from normal and Mises distributions, respectively [44-46] (Extended Data Fig. 1a). The insect tended to repeat its previous turning direction with the probability $P_{persist}$ (Extended Data Table 1) [45].

- The insect started following a wall when it was 2 cm or smaller near the wall [45]. The wall-following motion consisted of walking a distance of $L$ cm or stopping in $T_{stop}$ s, selected via $P_{stop}$ [45].

- The wall exit occurred with probability of $P_{exit}$ (Extended Data Table 1) [45]. The wall-departing angle β followed a log-normal distribution (Extended Data Fig. 2a, Extended Data Table 1) [44].

The Predictive Feedback Navigation Program and the insect's locomotory reaction to its cerci stimulation were embedded in the simulation environment to emulate controlled insects [2]. The reaction was experimentally collected with N = 4 insects, n = 80 trials (Extended Data Table 1).

Simulation and Evaluation of Stochastic Exploration Strategies

In Phase I, the stochastic exploration strategy would provide the cyborg insect with different destinations. After each arrival, a new destination was established. The stochastic exploration should distribute these destinations efficiently to maximize the acquired information. Herein, four stochastic exploration strategies were simulated. In these strategies, the relative orientation between the destination and the insect would be uniformly drawn from 0 deg to 360 deg [21]. Its distance to the insect would be governed differently by each strategy.

- Strategy 1: The distance was fixed as 0.5 m, the minimal effective working range of the human detection algorithm [2]. Thus, this strategy was named "Fixed Length."

- Strategy 2: The distance was drawn from a Levy distribution, which consecutively generated short distances and occasionally inputted a significantly long one, commonly discussed as an efficient stochastic exploration strategy [21,41]. The process of generating this distribution was inherited from [21]. This strategy was named "Levy Walk." The minimum distance was set as 0.5 m.

- Strategy 3: The distance was drawn uniformly from 0.5 m to 20 m, thus named "Uniform Distribution."

- Strategy 4: The strategy emulated the Brownian motion [47], i.e., the cyborg insect changed its destination whenever encountering walls. This process was done in a bounded terrain by fixing the distance as the terrain's longest length (30 m for this study). This strategy was named "Brownian Walk."

Herein, different searching mechanisms were studied. Fixed Length and Brownian Walk would favor local and global explorations, respectively. Levy Walk & Uniform Distribution might have a more balanced approach. The insect's natural motion was also simulated. The simulative arena was $20 \times 20$ m$^2$ (Extended Data Fig. 2c). The cyborg insect's initial coordinate was (-9.5, -9.5) m (Extended Data Fig. 2c). The simulative duration was 24 hours. Two simulation scenarios were simulated, with and without a searching target. The coverage percentage and the searching time were used for comparison. The cyborg insect was equipped with a target detection sensor having a circular working range of 0.5 m radius. When the system was 10 cm near the wall due to a distant destination, it would be first directed away from the wall. Then a new destination would be generated. The target's coordinate was (4.0, -4.5) m randomly selected (Extended Data Fig. 2c).

Demonstration Setup

*Indoor demonstration*

The exploration strategy was implemented onboard and used to search for a sitting human inside the region of $4.8 \times 6.6$ m$^2$. The cyborg insect was initially placed at the region's left corner. Its orientation faced was away from the human. The human was at the top right corner. The human location was unknown to the cyborg insect. The ambient temperature was 25 °C. The external optical tracking system (VICON®) monitored the hybrid system's position. The IR camera operated at the speed of 1 Hz.

*Outdoor Demonstration*

For the outdoor demonstration, which utilized a gait-adaptive IMU for localization, the experiment took place on the pavement surrounding our building, covering an area of $3.5 \times 6.0$ m². A human was seated at an intersection within the pavement area, making them invisible to the cyborg insect's camera unless it approached this intersection, located approximately 3 meters from the starting point. The ambient temperature was 28 °C. The cyborg insect was equipped with a backpack mounted IMU, providing yaw angle and position data, and the IR camera continued to operate at 1 Hz (Fig. 1c).

IMU data collection of free-walking insect

The acceleration and heading of the IMU on free walking insect were recorded at 100 Hz data rate. Cockroaches were left walking freely in 8 x 12 m arena. The data generated by Digital Motion Processor of the IMU (3D-Accelerations and Quaternions) was recorded to on-board flash memory during the free-walking procedure to obtain highest possible logging data rate provided by the system. Data was retrieved after the experiment for post-processing.

IMU-based localization system for cyborg insect system in 2D and 3D

First, the controller backpack was rested on the arena with its body-fixed coordinate system aligned with the coordinate system of the external optical tracking system or referencing coordinate system. The backpack was rotated around z, y and x axis so that yaw, roll and pitch value changed from varied in full rotation (full -180° to 180° scale for yaw and roll angle and -90° to 90° scale for pitch angle). The output Euler angle was compared with accurate orientation from the optical tracking system to make sure that their difference did not exceed 3.6° (1% of full-scale range). The backpack would be restarted and recalibrated if the previous condition did not meet requirement threshold. Next, the backpack was mounted on the back of the insect using double-sided tape. Then the cockroach was placed back to the origin of the referencing coordinate system and hold in place for a brief period. Afterward, experimenter released the cockroach for free walking for approximately 5 seconds. Distance travelled calculated by onboard IMU-based localization system and external optical tracking system was compared for gain-k recalibration by equation:

$$K_{adjusted} = \frac{Measured\ Travelled\ Distance\ by\ IMU}{Actual\ Travelled\ Distance\ by\ VICON} * K_{seed}$$

where $K_{seed}$ was set at 3.5.

The recalibration was required one time for each individual cockroach and would be used for all experiment with that cockroach afterward.

Once the gain k had been calibrated for the cockroach, it would be updated to the backpack controller wirelessly. Realignment of body-attached coordinate system and referencing coordinate system was re-applied and the cockroach was allowed to walk freely around the arena. Experiment would re-direct the moving direction of the cockroach by slightly tough the antenna of the cockroach if it approached the boundary of the tracking area of the optical tracking system to ensure minor data loss of referencing position. Positions of the insect was calculated on-board and streamed to center station GUI by BLE 5.1 to compare with referencing position.

For 3D investigation, cockroaches carrying backpack were guided to walk on a foam slope with incline angle of 7.9°. The insects were directed along the predetermined path and made sure not to fall off from the surface by experimenter. Position of the insects was calculated by the backpack controller and sent to center station GUI in real time by BLE 5.1 to compare with referencing position.

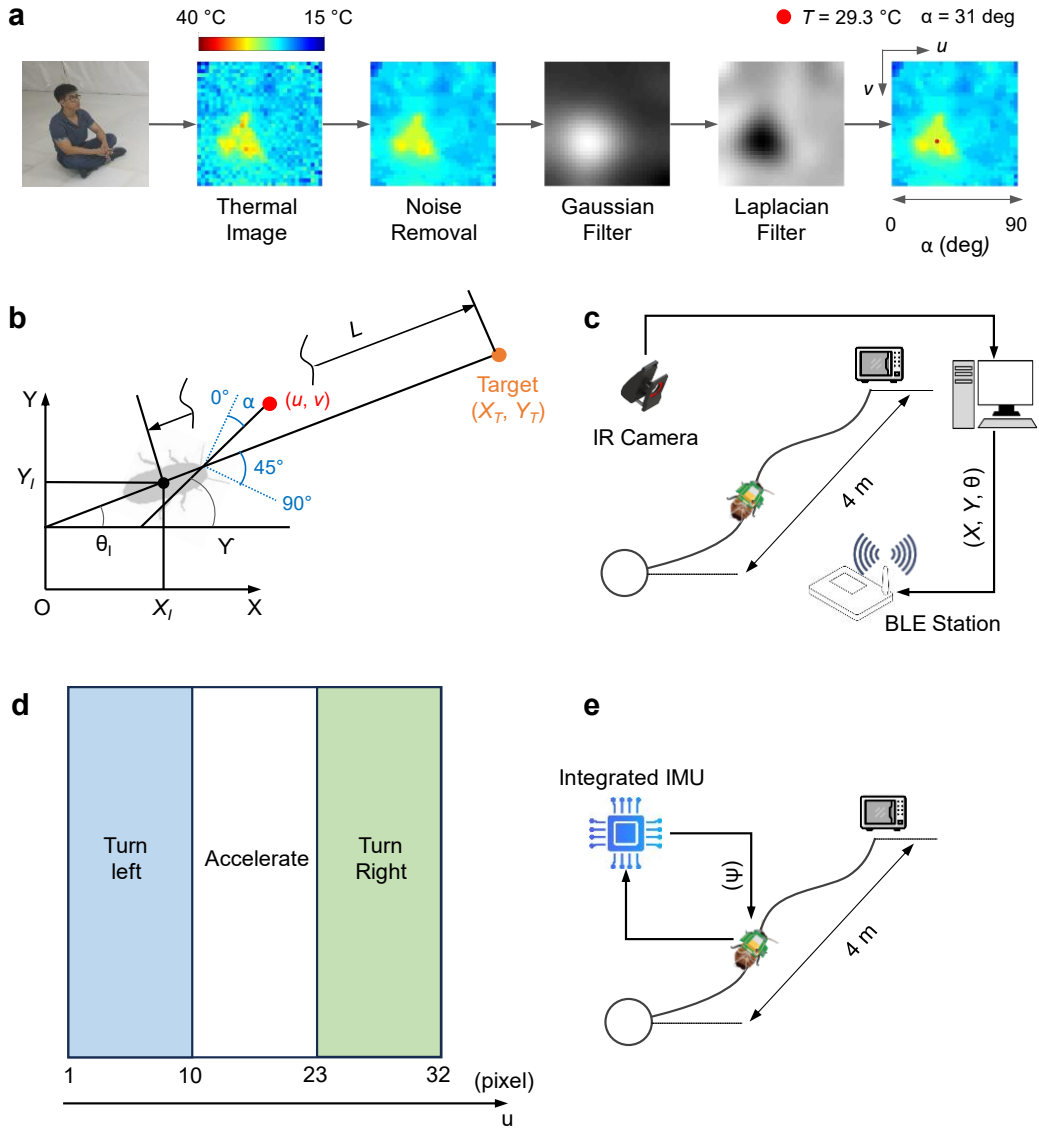

**Extended Data Fig. 1 | Thermal Source-Based Navigation Algorithm.** a, The blob detection technique. This technique uses multiple filters to locate thermal blobs inside the IR image. The red dot denotes the blob center. b, The target's position (XT, YT) is geometrically estimated using the insect's location (XI, YI, θ), the detected blob (i.e., the angle α), and the approaching step L. c, Experiment setup. The insect's coordinate is monitored, and wirelessly streamed to the onboard navigation program. The wireless communication is Bluetooth Low Energy (BLE). d, Navigation Logic: the decision-making process based on the blob center's u-axis position within the thermal image: the cyborg insect turns left if the position is within 1 to 10 pixels, turns right if within 23 to 32 pixels, and accelerates if within 11 to 22 pixels. e, Outlines the experiment setup for the cyborg insect, utilizing an integrated IMU to adjust its orientation based on the yaw angle (ψ), aiding the IR camera in capturing the heat source during the movement of the cyborg insect.

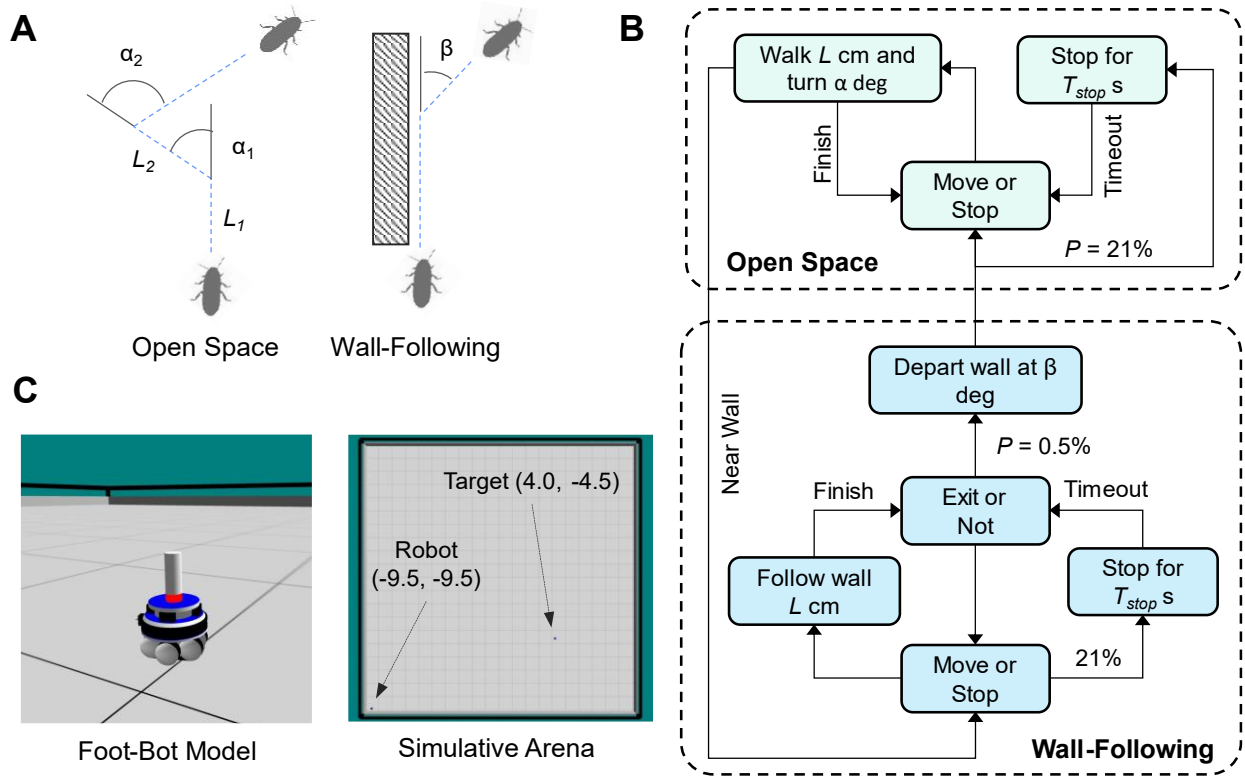

**Extended Data Fig. 2 | Design of the ROS simulation environment**. **a**, Model of an insect's 2D movement. Its motion in open space is segmented with straight walking step L and turning angle α. After following, it departs from the wall with an angle β. **b**, Finite-state machine modeling the insect's natural locomotion. The machine switches between the insect's various locomotory statuses covering two scenarios, i.e., the insect being in open space and near walls. **c**, Visualization of the simulation environment, ARGoS, and its mobile robot model, foot-bot. The robot is initially placed at the bottom left corner of the simulated terrain.

**Extended Data Table 1 | Parameters modelling the insect's 2D locomotion**

|  | Description | Value | Unit |
|---|---|---|---|
| $L$ | Straght walk distance | 17.5 ± 4.8 [45] | cm |
| $\alpha$ | Turning angle | Mises distribution [46] | deg |
| $P_{persist}$ | Probability of repeating turning direction | 71% [44] |  |
| $P_{stop}$ | Probability of stopping | 21% [45] |  |
| $T_{stop}$ | Stopping time | 36 ± 24 [45] | s |
| $P_{exit}$ | Probability to exit walls | 5% [45] |  |
| $\beta$ | Wall-departure angle | Log-normal [44] <br> $\mu$ = 36.6 <br> $\sigma$ = 2.1 | deg |
| $v$ | Linear speed | Natural: 2.0 ± 0.3 [45] <br> Stimulated: 8.4 ± 3.8 | cm/s |
| $\omega$ | Angular speed | Stimulated: 86.5 ± 43.5 | deg/s |